\documentclass[letterpaper]{article} 
\usepackage{aaai25}  
\usepackage{times}  
\usepackage{helvet}  
\usepackage{courier}  
\usepackage[hyphens]{url}  
\usepackage{graphicx} 
\usepackage{natbib}  
\usepackage{caption} 
\usepackage{algorithm}
\usepackage{algorithmic}
\usepackage{makecell}
\usepackage{mathtools}
\usepackage{amsfonts}
\usepackage{amsmath}
\usepackage{booktabs}
\usepackage{subfigure}
\usepackage{multirow}
\usepackage{multicol}
\usepackage{subcaption}
\usepackage{bbding}
\usepackage{pifont}
\usepackage[table,xcdraw]{xcolor}
\usepackage{subcaption}
\usepackage{newfloat}
\usepackage{listings}
\DeclareCaptionStyle{ruled}{labelfont=normalfont,labelsep=colon,strut=off} 
\floatstyle{ruled}
\newfloat{listing}{tb}{lst}{}
\floatname{listing}{Listing}
\makeatletter
\renewcommand*{\@fnsymbol}[1]{\ensuremath{\ifcase#1\or  \dagger\or \ddagger\or
   \mathsection\or \mathparagraph\or \|\or \or \dagger\dagger
   \or \ddagger\ddagger \else\@ctrerr\fi}}
\makeatother

\title{Pruning Large Language Models with Semi-Structural Adaptive Sparse Training}
\author{
    Weiyu Huang, Hu Yuezhou, Guohao Jian, Jun Zhu, Jianfei Chen\footnote{Corresponding Author.}\\
}
\affiliations{
    Dept. of Comp. Sci. and Tech., Institute for AI, BNRist Center, THBI Lab, \\

    Tsinghua-Bosch Joint ML Center, Tsinghua University \\

    \{hwy23, huyz21, jgh22\}@mails.tsinghua.edu.cn;\{dcszj, jianfeic\}@tsinghua.edu.cn
}

\usepackage{bibentry}

\begin{document}

\maketitle

\begin{abstract}
The remarkable success of Large Language Models (LLMs) relies heavily on their substantial scale, which poses significant challenges during model deployment in terms of latency and memory consumption. Recently, numerous studies have attempted to compress LLMs using one-shot pruning methods. However, these methods often suffer from considerable performance degradation on complex language understanding tasks, raising concerns about the feasibility of pruning in LLMs. To address this issue, we propose Adaptive Sparse Trainer (AST), a novel and efficient retraining framework tailored for semi-structured sparse models. AST enables models to learn optimal masks during the weight update process without incurring additional computational overhead. Furthermore, we demonstrate that incorporating knowledge distillation significantly improves retraining efficiency and enhances model performance under fixed computational constraints. Additionally, a supplementary set of well-initialized parameters is integrated to further augment the model's efficacy. AST achieves state-of-the-art performance with minimal training cost. When applied to the LLaMA2-7B model, AST reduces the perplexity and zero-shot accuracy gap between dense and 2:4 semi-structured sparse models to 0.6 and 1.16\%, respectively, utilizing less than 0.4\% of the pretraining tokens and GPU hours. Our work demonstrates the feasibility of deploying semi-structured sparse LLMs and offers a promising alternative for achieving highly compressed models when combined with existing quantization techniques. 
\begin{links}
\link{Code}{https://github.com/thu-ml/Adaptive-Sparse-Trainer}
\end{links}

\end{abstract}

%

\section{Introduction}

Transformer-based Large Language Models (LLMs) are equipped to handle complex tasks \cite{devlin2018bert, brown2020language, achiam2023gpt} and exhibit emergent abilities \cite{wei2022emergent} due to their expanding parameter count. However, this continual growth in model size poses significant challenges for practical deployment. In particular, inference speed suffers due to the increasing computational and memory demands. This has spurred a series of efforts to develop effective model compression techniques aimed at reducing memory footprints and easing the constraints associated with deploying these large-scale models.

Model pruning  \cite{frantar2023sparsegpt, han2015deep, sun2023simple} is an effective technique for compressing LLMs by setting a proportion of weights to zero, thereby adhering to a specific sparsity pattern. Recently, N:M sparsity has emerged as a type of semi-structured sparsity pattern that offers an optimal balance between precision and hardware efficiency. Specifically, N:M sparsity retains only N nonzero elements out of every group of M elements. This sparsity pattern can accelerate both matrix multiplication and memory access, potentially enhancing the performance of both pre-filling and decoding processes on off-the-shelf GPUs. Despite the promise of N:M sparsity, current state-of-the-art methods for pruning LLMs, such as SparseGPT \cite{frantar2023sparsegpt} and Wanda \cite{sun2023simple}, employ a post-training approach that determines the sparsity pattern in a layer-by-layer fashion without back-propagation. Although these methods improve efficiency, they can lead to significant performance degradation, particularly in knowledge-intensive tasks \cite{jaiswal2024compressingllmstruthrarely}, raising concerns about the feasibility of pruning LLMs. Additionally, while retraining pruned models has been successful in the pre-LLM era \cite{wang2019structured, lee2018snip, evci2020rigging}, its application to models with billions of parameters remains under-explored.

Although retraining sparse LLMs holds significant potential, it introduces several unique challenges: (1) Retraining is computationally expensive, necessitating techniques that ensure rapid convergence. (2) The output model must adhere to strict sparsity pattern, which adds additional constraints to the retraining process. (3) In order to achieve optimal performance, both masks and weights should be learnable during training. (4) Pruning LLMs may compromise essential language understanding and reasoning abilities, which are significantly harder to restore compared to simpler metrics such as perplexity. To this end, we propose a novel and efficient training method, Adaptive Sparse Trainer (AST), designed to produce high-performance sparse LLMs. \textit{For the first time, we demonstrate that pretrained 2:4 sparse LLMs can achieve competitive performance not only in terms of perplexity but also in more demanding knowledge-intensive tasks}, making them viable for practical deployment. AST integrates the transition from dense to sparse models directly into the training process, gradually decaying unimportant weights to zero through a carefully designed decay mechanism. This approach allows for the revival of pruned weights during training, enabling a smooth and dynamic adjustment of the model's connectivity pattern while adhering to the N:M sparsity structure. Furthermore, AST applies knowledge distillation using dense model as the teacher, which can significantly speed up model convergence and prevent sparse models from settling into local optima. It also helps the sparse model retain the valuable world knowledge and performance characteristics of the original dense model, thereby enhancing its generalization ability and compensating for the use of weaker training datasets. To further enhance performance, a supplementary set of parameters is integrated using information from pruned weights. When applied to the LLaMA2-7B model, AST achieves a 2:4 sparse configuration with minimal performance loss, demonstrating that compressed models can still perform effectively in practice. Our model further benefits from AWQ quantization, achieving competitive compression rates with state-of-the-art performance.

The key contributions of this work are as follows:

\begin{itemize}
\item We propose Adaptive Sparse Trainer (AST), an novel semi-structured pruning framework designed to compress large language models efficiently while maintaining high performance.
\item AST introduces a gradual decay scheduler (Annealing SR-STE) combined with knowledge distillation to accelerate model convergence and improve performance on complex tasks.
\item We introduce SLoRB, a technique that boosts model performance by adding a supplementary set of well-initialized parameters.
\item When applied to LLaMA2-7B model, our 2:4 sparse model experiences only a 0.6 increase in perplexity and a 1.16\% accuracy drop in zero-shot tasks, with less than 0.4\% of the pretraining cost.
\end{itemize}

\section{Related Work}

\paragraph{Network Pruning}
Pruning is a prevalent technique aimed at reducing model size and computational costs. It originated from methods like OBD \cite{lecun1989optimal} and OBS \cite{hassibi1993optimal}. Based on sparsity patterns, pruning methods can be broadly categorized into unstructured, structured, and semi-structured pruning.
Unstructured pruning removes individual weights \cite{han2015deep, paul2022unmasking}, which can maintain performance even with high sparsity. However, due to its random pattern, unstructured models are difficult to accelerate. Structured pruning \cite{liu2017learning, molchanov2019importance, nova2023gradient, shen2022structural}, on the other hand, removes entire neurons, filters, or attention heads, resulting in models that are easier to accelerate on standard hardware but often suffer from severe performance loss. Semi-structured sparsity (e.g., N:M sparsity) \cite{hubara2021accelerated} has been applied as a trade-off between performance and achieving actual speedup. Recently, a series of works \cite{frantar2023sparsegpt, sun2023simple, zhang2024plug, zhang2023dynamic} have made progress in pruning LLMs with billions of parameters. However, these pruned models from training-free methods still fall short in complex zero-shot performance.

\paragraph{Retraining Pruned Models}

Another line of research \cite{NIPS2015_ae0eb3ee, NEURIPS2020_d1ff1ec8, renda2020comparing, zhou2023three} has focused on retraining pruned models to enhance performance. While retraining has been shown to be effective with smaller models and simpler tasks \cite{kurtic2022gmp, zhu2017prune}, it often involves additional training steps \cite{frankle2018lottery} or introduces extra parameters for the pruning process \cite{shi2023upop}, which limits its application to large-scale models due to the substantial computational resources required. Other retraining methods focus on unstructured sparsity, which may struggle when given stricter sparsity patterns \cite{lee2018snip, evci2020rigging}. Recently, Sheared LLaMA \cite{xia2023sheared} employed a two-stage structured pruning process to train models that outperform others of similar size. However, they employ a computationally intensive pruning stage, unsuitable for more fine-grained sparsity. Our work proposes a lightweight pruning process that enables the retraining of semi-structured sparse large language models with minimal training costs.

\paragraph{Combining Pruning with Quantization} 
Pruned models can be further compressed using quantization. Earlier methods like Deep Compression \cite{han2015deep} combined pruning and quantization to significantly reduce the size of neural networks. More recent work \cite{frantar2023sparsegpt, sun2023simple, guo2024compressing} has combined sparsity with quantization in large language models. In our work, we report results using AWQ quantization \cite{lin2023awq} with our semi-structured sparse model.
\section{Methods}

\begin{figure*}[t]
    \centering
    \includegraphics[width=18cm,height=4.5cm]{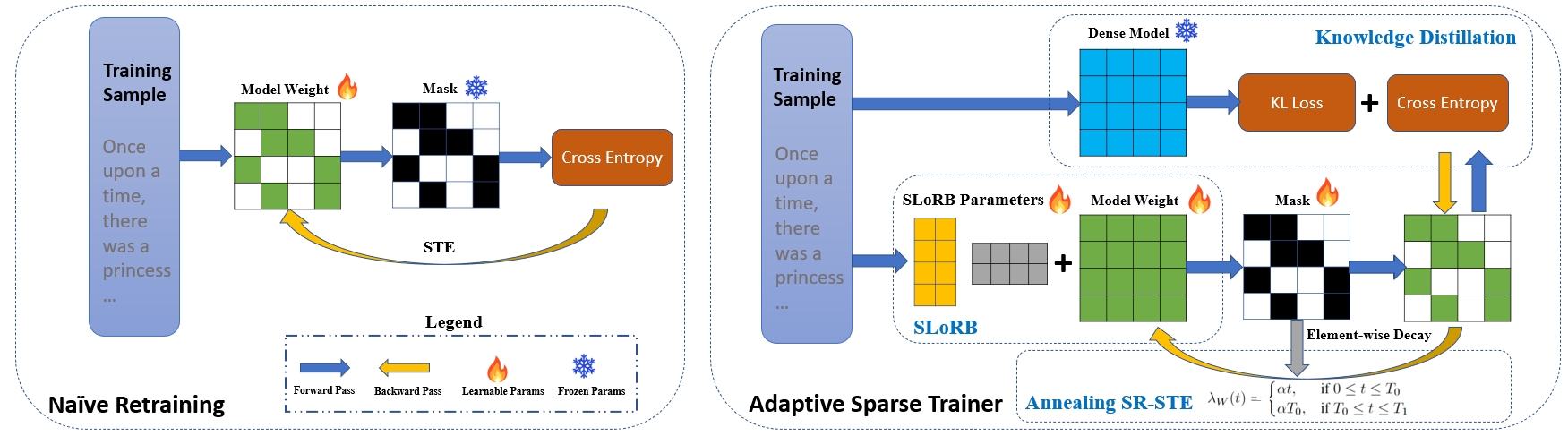}
    \caption{\textbf{(Left)} In the naive training baseline, the mask remains constant during training, which can result in suboptimal performance. \textbf{(Right)} Adaptive sparse training enables both mask and weight learning through a scheduled decay term. AST also utilizes distillation and SLoRB parameters to speed up convergence and improve performance.} 
    \label{fig:1}
\end{figure*}

We begin by revisiting a naive approach to training sparse models before introducing our method. The full training pipeline is illustrated in Figure \ref{fig:1}.

\subsection{Sparse Training}\label{4.1}
Pruning model weights is equivalent to multiplying them by an element-wise mask. For example, consider the matrix multiplication in a linear layer:
\begin{align}
    Z = XW^\top, \quad Z \in \mathbb{R}^{N \times D}, \quad X \in \mathbb{R}^{N \times C}, \quad W \in \mathbb{R}^{D \times C}, \nonumber
\end{align}
where $X$, $W$, and $Z$ represent the model input, weight matrix, and output activation, respectively. The pruned weight matrix can be expressed as:
\begin{align}
    \tilde{W} = m(W) \odot W, \quad m(W) \in \{0,1\}^{D \times C}, \nonumber
\end{align}
where $m(\cdot)$ is a mapping that selects a mask based on the weight matrix $W$. In this work, we focus on N:M sparsity \cite{hubara2021accelerated}, where there are $N$ nonzero elements in every $M$ consecutive weights in the same row. However, when implementing the backward pass for a sparse model with automatic differentiation, the gradient cannot flow through the discrete mask $m(W)$. A straightforward approach to solve this issue is the straight-through estimator (STE) \cite{bengio2013estimating} which updates the parameters using the gradient with respect to the masked weight $\tilde{W_t}$, where $W_t$ is the parameter at iteration $t$:
\begin{align}\label{eq1}
    W_{t+1} \leftarrow W_{t} - \gamma_t g(\tilde{W_t}), 
\end{align}
here $g(\tilde{W_t})$ represents the gradient with respect to $\tilde{W_t}$ and $\gamma_t$ indicates the learning rate at iteration $t$.

A previous method \cite{sun2023simple} employs a fixed mask derived from one-shot pruning techniques and updates only the remaining parameters using language modeling loss. However, this strategy has two significant limitations. First, it discards the valuable information embedded in the pruned weights and prevents a smooth transition from a dense to a sparse model, resulting in slower convergence and suboptimal performance. Second, relying solely on the language modeling loss during retraining increases the risk of the sparse model becoming trapped in a local optimum, leading to poor outcomes.

To address these issues, we maintain all model weights and select the mask on-the-fly while gradually decaying non-essential weights to zero, thereby enabling a smooth transition ($\mathsection$\ref{4.2}). Additionally, we leverage knowledge distillation to guide the model toward a global optimum, significantly accelerating convergence under a fixed computational budget ($\mathsection$\ref{4.3}). Finally, we offer an optional method to further enhance model performance by incorporating an additional set of well-initialized parameters, as detailed in ($\mathsection$\ref{4.4}).

\subsection{Adaptive Mask Learning with Annealing SR-STE}\label{4.2}

As highlighted in previous work \cite{frankle2018lottery, evci2020rigging}, the connectivity pattern is as crucial as the weights themselves in determining the performance of sparse models. Therefore, we enable the model to adaptively learn the optimal mask during training instead of fixing it permanently. Specifically, we recalculate \( m(W_t) \) based on the magnitude criterion every \(\Delta t\) iterations and add a decay term to the masked weights. The intuition behind this approach is straightforward: important weights typically receive strong gradients of the same sign during training, causing their magnitude to grow despite the decay, and they will eventually be revived. In contrast, less important weights tend to receive weaker, mixed gradient signals, leading them to decay to zero and remain masked under our settings. We build upon SR-STE \cite{zhou2021learning} by applying L2 decay to masked weights, as this approach helps preserve overall model performance.

However, selecting the appropriate decay strength is challenging. If the decay signal is too strong, masked weights will struggle to regenerate, leading to an almost fixed mask. On the other hand, if the decay signal is too weak, the model will fail to filter out important weights, resulting in heavy oscillation that hinders model convergence. To address this dilemma between stability and exploration, we propose adding a moving decay schedule, with weaker decay strength at the beginning to encourage the model to explore various connectivity patterns and a stronger decay signal toward the end to facilitate model convergence.

Formally, we update the model weights with the following equation:
\begin{align} \label{eq2}
        \lambda_W(t)= 
\begin{cases}
    \alpha t, & \text{if } 0 \leq t \leq T_0, \\
    \alpha T_0, & \text{if } T_0 \leq t \leq T_1, 
\end{cases}
\end{align}
\begin{align} 
    W_{t+1} \leftarrow W_{t} - \gamma_t \left( g(\tilde{W_t}) + \lambda_W(t) \left( \overline{m(W_t)} \odot W_t \right) \right), \label{eq3}
\end{align}
where $\lambda_W(t)$ denotes the decay factor, $\overline{m(W_t)}$ denotes the mask for pruned weights at iteration $t$, $T_1$ denotes the total number of training batches, and $T_0$ represents the batch at which the decay factor stops increasing. Compared with STE in Equation \ref{eq1}, masked weights receive a decay signal proportional to the decay factor $\lambda_W(t)$, which starts small at the beginning and increases later on. This method, termed Annealing SR-STE (ARS-STE), has been shown to improve model performance compared to naive SR-STE. We also provide a detailed analysis of mask oscillation behavior during training in Section 7 of the Appendix.

We experimented with various one-shot pruning criteria incorporating activation information \cite{frantar2023sparsegpt, zhang2024plug} for mask selection. However, the \emph{classic magnitude criterion consistently outperformed these methods in both computational efficiency and performance.} As activation-based approaches incur higher computational costs and are less reliable due to activation fluctuations caused by weight updates, leading to degraded performance in later training stages.

Since the naive SR-STE is only compatible with the SGD optimizer, we introduce further adjustments to support more advanced optimizers such as AdamW. Specifically, as AdamW maintains both first-order and second-order moments of the gradient, we decouple the weight decay term from the first-order moment instead of directly adding decay to the gradient, as done in \cite{hu2024accelerating}. This separation ensures that the weight decay term depends solely on the current weight, avoiding interference with momentum calculations. We then use the decoupled first-order signal to update the second-order moment. Detailed expressions and explanations are provided in Section 6 of the Appendix.

\subsection{Alleviating the Retraining Dilemma through Knowledge Distillation} \label{4.3} 

A key distinction between pretraining and retraining pruned models lies in how their parameters are initialized. In the pretraining setting, parameters are typically randomly sampled, whereas in retraining, they are inherited from a well-trained model. Consequently, pruned models retain a portion of the patterns acquired during pretraining.

\begin{figure}[t]
    \centering
    \includegraphics[width=0.4\textwidth]{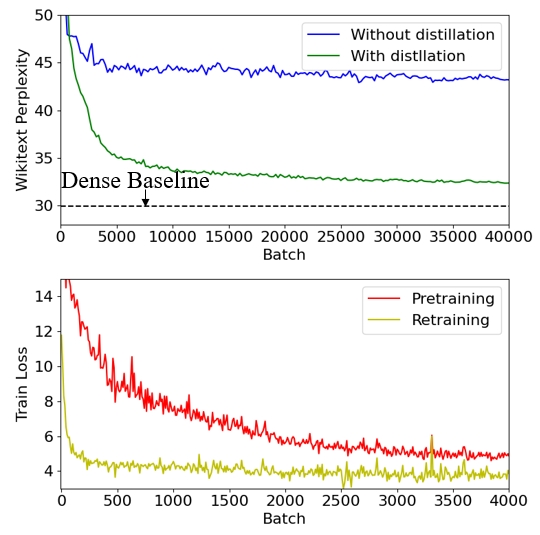}
    \caption{(\textbf{Upper}) The Wikitext perplexity curve for retraining GPT2 with and without knowledge distillation. (\textbf{Lower}) The training loss curve for pretraining and retraining.}
    \label{pic2}
\end{figure}

We discovered that this feature tends to trap pruned models in local optima. Specifically, while retraining can achieve much faster initial convergence, it often fails to reach an optimal state later on; we refer to this phenomenon as the \textbf{Retraining Dilemma}. As illustrated in Figure~\ref{pic2}, although using cross-entropy loss during retraining initially leads to a quicker reduction in training loss compared to pretraining, the test set perplexity remains unstable and elevated. We attribute this to the fact that, unlike randomly initialized models, the pruned model is more prone to overfitting on the limited data available at the start due to its prior knowledge. This overfitting hampers the model’s ability to learn global features and achieve optimal convergence. Notably, this issue persists even when using a significantly smaller learning rate than in pretraining, suggesting that simply reducing the learning rate is insufficient to resolve the problem.

As a solution, we found that applying KL-divergence loss \cite{kullback1951information} can address this issue. Unlike the language modeling loss, KL-divergence loss measures the difference between two probability distributions, offering richer signals that help mitigate overfitting in the early stages of training. Consequently, we employ the following loss function:
 \begin{align}
    \mathcal{L}_{logit} = D_{KL}(p_{\theta_t} || p_{\theta_s}) &= \frac{1}{B \times S} \sum_{i=1}^{B \times \text{seq}} p_{\theta_t}(x_i) \log \frac{p_{\theta_t}(x_i)}{p_{\theta_s}(x_i)}, \nonumber 
 \end{align}
 \begin{align} \label{eq4}
     \mathcal{L} =& \alpha \mathcal{L}_{logit} + (1-\alpha)\mathcal{L}_{task},   
 \end{align}
where $\mathcal{L}_{task}$ is the cross-entropy loss, $B$ is the batch size, $S$ is the sequence length, and $p_{\theta_t}$ and $p_{\theta_s}$ are the probability distributions of the teacher and student models, respectively.

We observe that the Retraining Dilemma is more pronounced in smaller models; therefore, we typically apply a larger $\alpha$ for these models. We also experimented with various distillation methods that utilize intermediate information for distillation. However, we found that adding constraints to intermediate outputs hinders the model's generalization ability and leads to undesirable outcomes.

\subsection{Sparse Low-Rank Boosting}\label{4.4}

The pruned model with 2:4 sparsity, which retains only half of the parameters under strict structural constraints, experiences a reduction in expressive capacity. To mitigate this, we incorporate a LoRA (Low-Rank Adaptation) \cite{hu2021lora} shape parameter that is trained alongside the original parameters, helping to bridge the performance gap between dense and sparse models with minimal memory overhead. Unlike traditional LoRA fine-tuning, where the original model parameters are frozen and only the adapters are trained, our approach trains both the sparse parameters and the adapter weights simultaneously. This approach recognizes that while the classic LoRA method freezes the original model to prevent overfitting and catastrophic forgetting during downstream tasks, our method enhances generalization by training both the additional and original weights on a pretraining dataset, thereby enabling concurrent training.

Another notable aspect of our method is the initialization strategy. Classic LoRA typically employs random initialization techniques, such as Xavier initialization \cite{glorot2010understanding} or Kaiming initialization \cite{he2015delving}. However, in our approach, we leverage the pruned weights as additional information to initialize the adapter weights, thereby accelerating the training process. Given that in 2:4 sparsity, each neuron can only access half of the inputs, we find that incorporating additional information to retain the weight's first-order information helps preserve the model's capacity. Specifically, for a weight matrix \( W \) of size \( N \times d \) and its corresponding mask matrix \( M \), we select the rank \( r \) as \( \frac{d}{k} \), where \( k \) can be 64, 32, 16, and so on. A smaller \( k \) improves performance but also increases memory usage. We select a projection matrix \( X \) of size \( \frac{d}{k} \times d \) and a SLoRB weight matrix \( S \) of size \( N \times \frac{d}{k} \). The matrices \( X \) and \( S \) are defined as follows:
\begin{align} \label{eq5}
    x_{ij} = \begin{cases}
    1,& \text{if } i \cdot k \leq j \leq (i+1) \cdot k - 1,\\
    0,              & \text{otherwise},
\end{cases}
\end{align}
\begin{align} \label{eq6}
    S_{ij} = \frac{1}{k} \sum_{p=j \cdot k}^{ (j+1) \cdot k -1} W_{ip}  \cdot {\neg M_{ip}}.\ 
\end{align}

\begin{figure}[t]
  \centering \includegraphics[width=0.49\textwidth,height=2.5cm]{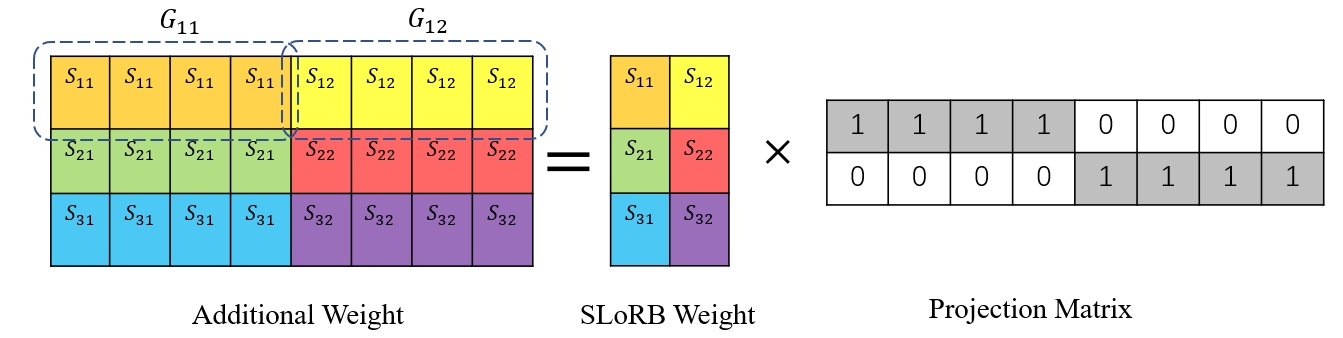}
  \caption{
  \textbf{Visualization of SLoRB Initialization}: Consider a weight matrix of size 3 by 8, with $k=4$. Weight $S_{ij}$ is broadcasted within group $G_{ij}$ after multiplication with the projection matrix.}
  \label{pic3}
\end{figure}

We define Group \( G_{ij} \) as the elements from \( j \cdot k \) to \( (i+1) \cdot k - 1 \) in row \( i \). As illustrated in Fig \ref{pic3}, each SLoRB weight \( S_{ij} \) is broadcast within Group \( G_{ij} \). By setting \( S_{ij} \) as the mean of all pruned weights in Group \( G_{ij} \), this design ensures that the total mean of groups \( G_{ij} \) remains consistent after pruning. Although different initialization methods may ultimately converge to similar outcomes given sufficient training steps, our method converges more rapidly, making it particularly advantageous when the computational budget is limited, as demonstrated in Section 4 of the Appendix. We refer to this method as \textbf{Sparse Low-Rank Boosting (SLoRB)}, which is an optional technique that trades off memory overhead for enhanced performance. The complete pseudo-code for our method is provided in Algorithm \ref{alg:algorithm}.

\begin{algorithm}[h]
\caption{Training Process for AST}
\label{alg:algorithm}
\textbf{Input}: Weight of linear layers $\mathcal{W}^{(0)}$; Mask update frequency $\Delta t$; Total training iteration $T_0$;  SLoRB parameter k; Decay increase iteration $T_1$; 
\begin{algorithmic}[1] 
\FOR{$W^{(0)} \in \mathcal{W}^{(0)}$}
\STATE Initialize the mask {$m(W^{(0)})$} based on magnitude;
\IF {Use SLoRB}
\STATE Initialize SLoRB weight $S^{(0)} \in \mathcal{S}^{(0)}$ and $X^{(0)} \in \mathcal{X}^{(0)}$ by Equation \ref{eq5} and \ref{eq6};
\ENDIF
\ENDFOR

\FOR{t = 1,2,..$T_0$}
\IF{t$\mod$$\Delta t$ ==0}
\FOR{$W^{(t)} \in \mathcal{W}^{(t)}$}
\STATE Update model mask mask {$m(W^{(t)})$};
\ENDFOR
\ENDIF
\STATE Compute decay for term $\lambda_W(t)$ from Equation \ref{eq2}; 
\STATE Compute gradient $g(W^{(t)})$ by back-propagation on distillation loss $\mathcal{L}$ from Equation \ref{eq4};
\STATE Update weight $W^{(t)}$ by Equation \ref{eq3};
\IF {Use SLoRB}
\STATE Update $S^{(t)}$ and $X^{(t)} $ by gradient;
\ENDIF
\ENDFOR
\STATE \textbf{return} the pruned model.
\end{algorithmic}
\end{algorithm}

\section{Experiments}
\begin{table*}[ht]
\renewcommand{\arraystretch}{0.2}
  \caption{Perplexity results on raw-WikiText2 on 2:4 sparsified language models. AST outperforms both training-free and other training-based methods with similar computational costs.}
  \label{tab1}
  \centering
\resizebox{0.85\textwidth}{!}{%
\begin{tabular}{ccccccccc}
\toprule
\multicolumn{1}{l}{} &
  \multicolumn{1}{l}{} &
  \multicolumn{3}{c}{OPT} &
  \multicolumn{4}{c}{GPT2} \\ \cline{3-9} 
Method &
  \makecell{Training} &
  \multicolumn{1}{c}{125M} &
  \multicolumn{1}{c}{350M} &
  \multicolumn{1}{c}{1.3B} &
  124M &
  350M &
  774M &
  1.5B \\ \midrule
Dense    & -  & 27.76 & 22.00 & 14.62 & 29.95 & 21.72 & 19.43 & 17.40 \\ \midrule
SparseGPT         &  \XSolidBrush  & 45.58 &    40.33   &  29.03  & 50.09 & 31.03  & 25.98 & 21.14  \\
\rowcolor[HTML]{C0C0C0}
Wanda  &  \XSolidBrush  &  60.91  & 50.16 & 23.92 &  115.64  & 63.71  & 49.97  &  30.44  \\
Iterative Magnitude Pruning    &  \CheckmarkBold  &38.37 &30.29 & 23.94 & 40.08 & 29.86  & 24.31 & 20.83  \\
\rowcolor[HTML]{C0C0C0}
Gradual Magnitude Pruning  &  \CheckmarkBold  & 31.51 & 25.98&  16.78  & 33.48 & 24.83& 22.01 & 18.96 \\ 
{ AST-Naive(Ours)} & \CheckmarkBold  & 30.22
   & 24.65  & 15.85   &   32.23 &  23.65 & 21.29&
  18.33 \\
\rowcolor[HTML]{C0C0C0}
AST-Boosted(Ours)     & \CheckmarkBold   & \textbf{28.68} & \textbf{24.03}  &  \textbf{15.43} &    \textbf{31.13}   & \textbf{23.03} &  \textbf{20.66} & \textbf{18.01} \\ \bottomrule
\end{tabular} %
}
\end{table*}
\begin{table*}[ht]
\centering
\caption{Accuracy (\%) of various open-sourced models on seven zero-shot tasks.}
\resizebox{1.0\textwidth}{!}{%
\begin{tabular}{cccccccccc}
\toprule
\textbf{Models \tiny{(Sparsity Pattern)}} &
\textbf{Parameters} &
  \textbf{BoolQ} &
  \textbf{RTE} &
  \textbf{HellaSwag} &
  \textbf{WinoGrande} &
  \textbf{ARC-e} &
  \textbf{ARC-c} &
  \textbf{OBQA} &
  \textbf{Mean} \\ \midrule
  LLaMA2-7B \tiny{(Dense)}& 6.7B   & 77.73 & 63.89 & 57.18 & 69.04 & 76.17 & 42.91 & 31.60 & 59.78 \\ \midrule
  {Sheared-LLaMA-2.7B \tiny{(Dense)}} & 2.7B  & 65.99 & 50.54 & 51.21 & 65.03 & 67.29 &33.27 & 28.80 & 51.73 \\ 
  {INCITE-Base-3B \tiny{(Dense)}} & 2.8B  & 67.40 & 52.34 & 47.91 & 62.98 & 67.68 & 31.74 & 27.60 & 51.09 \\ 
{Open-LLaMA-3B-v2 \tiny{(Dense)}} & 3.4B  & 65.89 & 55.69 & 52.12 & 63.59 & 68.34 & 34.32 & 26.00 & 52.13 \\ 
{LLaMA-7B \tiny{(Sparse)}} & 3.4B  & 73.21 & 61.34 & 54.86 & 66.18 & 70.24 & 35.68 & \textbf{31.80} & 56.19 \\ \midrule
LLaMA2-7B \tiny{(Sparse)}& 3.4B   & 73.12 & \textbf{66.06} & 54.66 & 67.87 & 73.61 & 39.93 & 28.60 & 57.68 \\
LLaMA2-7B \tiny{(Sparse+SLoRB)} & 4.2B    & \textbf{75.04} & \textbf{66.06} & \textbf{55.24} & \textbf{68.48} & \textbf{74.91} & \textbf{41.11} & 29.40 & \textbf{58.62} \\ 
 \bottomrule
\end{tabular}%
}
\label{tab2}
\end{table*}

\subsection{Experiment Setup}

\textbf{Model Configurations.} We report the performance of Adaptive Sparse Trainer (AST) across three different LLM model families: LLaMA2-7B \cite{touvron2023llama}, OPT \cite{zhang2022opt}, and GPT2 \cite{brown2020language}. We present results for two different sparsity patterns: AST-Naive, which adheres to a strict 2:4 sparsity pattern without additional weights, and AST-Boosted, which incorporates extra SLoRB weights. For AST-Boosted models, we selected $k=16$, introducing an additional 12.5\% of parameters. Optimal hyperparameters were identified through a grid search, with the specific hyperparameters and training details provided in Section 3 of the Appendix.

\textbf{Data.} For training smaller models like the OPT and GPT2 model families, we utilized the C4 dataset \cite{raffel2020exploring}. For the LLaMA2-7B model, we employed a more comprehensive dataset, RedPajama-v1\footnote{https://huggingface.co/datasets/togethercomputer/RedPajama-Data-1T}, which encompasses data from seven domains: CommonCrawl, C4, GitHub, Wikipedia, Books, ArXiv, and StackExchange. Additionally, we leveraged the dynamic batching feature provided in the ShearedLLaMA \cite{xia2023sheared} codebase.

\textbf{Baseline.} For the GPT2 and OPT models, we compare our methods with both training-free and training-based approaches. Among training-free methods, we include comparisons with SparseGPT \cite{frantar2023sparsegpt} and Wanda \cite{sun2023simple}. For training-based methods, given the lack of existing results for retraining generative language models with N:M sparsity, we implemented iterative magnitude pruning \cite{frankle2018lottery} and gradual magnitude pruning \cite{kurtic2022gmp}, both of which can be adapted to achieve the desired sparsity. To ensure a fair comparison in terms of computational cost, we report results for training-based methods that include distillation, as we have observed that incorporating distillation significantly enhances performance. Due to computational constraints, we report the results of training-based baselines only for the GPT2 and OPT models.

For the LLaMA2-7B model, we compare our approach against strong dense pre-trained LLMs with a similar number of non-zero parameters. Additionally, we include a comparison with the LLaMA-7B 2:4 sparsity model provided in the Wanda \cite{sun2023simple} paper.

\textbf{Evaluation.} We evaluated the WikiText-2 perplexity for all models and assessed zero-shot and few-shot performance on LLaMA2-7B using EleutherAI’s LM Harness \cite{gao2021framework}. Our evaluation tasks included zero-shot ARC Easy \cite{clark2018think}, OpenBookQA \cite{mihaylov2018can}, WinoGrande \cite{sakaguchi2021winogrande}, RTE from the GLUE benchmark \cite{wang2018glue}, HellaSwag \cite{zellers2019hellaswag}, ARC Challenge \cite{clark2018think}, BoolQ \cite{DBLP:journals/corr/abs-1905-10044}, as well as more complex tasks like the MMLU benchmark \cite{hendrycks2020measuring}, GSM8K \cite{cobbe2021training}, and MATH \cite{hendrycks2021measuringmathematicalproblemsolving}.

\subsection{Main Results}
\textbf{Language Modeling} In Table \ref{tab1}, we compare the perplexity results of our sparse models with those of the baselines. Our methods significantly outperform both training-free and training-based approaches across various models, achieving results close to those of the dense baseline. A plausible reason why training-based methods like IMP fail is that while a lottery ticket may exist for a randomly initialized model, this may not be the case for a well-trained model. As for the gradual baseline, applying a proportion of training steps at low sparsity may not provide the model with sufficient time to converge.

\noindent \textbf{Zero-shot Result} In Table \ref{tab2}, we present the accuracy results of seven zero-shot tasks for AST and baseline models. Our method performed best in most of the tasks. It should be noted that our sparse models consistently outperform smaller dense models with a similar parameter count, suggesting a promising approach for obtaining parameter-efficient models. Moreover, our method requires only a minimal amount of training tokens to converge. For example, when training the LLaMA2-7B model, \textbf{we utilized only 7B tokens, which is less than 0.4\% of those used in pretraining, making AST applicable to open-source models.}

Our method can also be seamlessly adapted to current quantization techniques to achieve extremely compressed models with minimal performance drop. We provide results using AWQ quantization methods in section 8 of the Appendix.

\subsection{Additional Results for LLaMA2-7B}

Recent findings \cite{jaiswal2024compressingllmstruthrarely} have shown that previous pruning methods suffer significant performance degradation in knowledge-intensive tasks. To address this, we tested our LLaMA2-7B model on more complex tasks. As shown in Table \ref{tab3}, our model preserves most of the knowledge from the dense model and continues to perform well in knowledge-intensive tasks compared with previous pruning methods. This provides strong evidence that 2:4 pruned models possess considerable potential, contrary to the observations in previous studies.

\begin{table}[ht]
\caption{Results of perplexity and knowledge-intensive tasks for LLaMA2-7B models with 2:4 sparsity. The symbols $\downarrow$ and $\uparrow$ indicate that lower and higher values are better, respectively. (LoRA methods are finetuned with $r$=64)}
\centering
\resizebox{0.5\textwidth}{!}{%
\renewcommand{\arraystretch}{1}
\begin{tabular}{ccccc}
\toprule
\textbf{LLaMA2-7B}    & Dense  & Wanda & AST-Naive & AST-Boosted \\ \midrule
\textbf{Perplexity} $\downarrow$ & 5.12   & 11.02  & 5.82      & \textbf{5.69} \\
\textbf{MMLU \tiny{(5-shot)}} $\uparrow$ & 45.3        & 27.6  & 37.9      & \textbf{38.2} \\
\textbf{MATH \tiny{(4-shot)}} $\uparrow$ & 5.38   & 2.86  & 4.42  & \textbf{4.64} \\
\textbf{GMS8K \tiny{(LoRA Finetuned)}} $\uparrow$ & 40.3   & 32.1  & 35.6      & \textbf{36.2} \\
\bottomrule
\end{tabular}
}
\label{tab3}
\end{table}

\subsection{Ablation Study}
\textbf{Training Sparse Models.} Our AST-Naive method employed distillation, Annealing SR-STE, and adaptive masking during training. In Table \ref{tab4}, we present an ablation study to assess the effects of each of these components. For a fair comparison, we apply additional training steps to non-distillation methods. Specifically, we analyze the impact of training without distillation, using a naive SR-STE decay factor, and fixing the mask during training, each component in isolation. Additionally, we provide results for the naive training baseline in Figure \ref{fig:1} mentioned above. We demonstrate that all three approaches contribute to performance gains compared to naive training across various models.

\begin{table}[h]
\centering
\caption{Ablation study of different methods in training sparse models.}
\resizebox{0.45\textwidth}{!}{%
\begin{tabular}{ccccc}
\toprule
 & \multicolumn{3}{c}{GPT} & OPT \\ \cline{2-5} 
 Method    & 124M & 350M& 774M &125M \\ \midrule
  Dense   & 29.95 & 21.72 & 19.43 &27.76   \\ \midrule
\rowcolor[HTML]{C0C0C0} Naive Training & 40.34 & 29.79 & 28.65 & 39.46     \\ 
  No distillation & 39.29& 29.08& 27.21 &36.97      \\
  \rowcolor[HTML]{C0C0C0}Static SR-STE    & 32.84 & 24.04 & 21.73 &  31.08       \\ 
 Fixed Mask    & 32.93 & 24.18& 21.95&   31.06  \\ 
 \rowcolor[HTML]{C0C0C0} AST-Naive(Ours)   & \textbf{32.23} & \textbf{23.65}& \textbf{21.29} & \textbf{30.22}  \\
\bottomrule
\end{tabular}
}
\label{tab4}
\end{table}

\noindent \textbf{Distillation Functions} We also conducted ablation studies on different distillation functions used in previous work, including TinyBERT \cite{jiao2019tinybert}, MobileBERT \cite{sun2020mobilebert}, and Sparse-Finetuning \cite{kurtic2023sparse}, which use attention and hidden states for distillation. Additionally, we examined MiniLLM \cite{gu2024minillm}, which employs reverse KL divergence. We find that using intermediate information during the distillation of generative language models is detrimental, and that KL loss is sufficient for optimal performance. Reverse-KL yields performance similar to that of forward KL. Detailed descriptions of the distillation loss functions for each of these methods are provided in Section 5 of the Appendix.

\begin{table}[t]
\centering
\caption{Ablation study on different distillation loss for training sparse models.}
\resizebox{0.45\textwidth}{!}{%
\begin{tabular}{ccccc}
\toprule
 & \multicolumn{3}{c}{GPT} & OPT \\ \cline{2-5} 
 Method    & 124M & 350M& 774M& 125M \\ \midrule
Dense   & 29.95 & 21.72   &19.43  &27.76\\  \midrule
\rowcolor[HTML]{C0C0C0}TinyBERT         & 42.75  & 33.35  & 30.86 & 39.39\\ 
MobileBERT       & 44.87 &   31.67  &  29.75 & 40.33  \\ 
\rowcolor[HTML]{C0C0C0}Sparse-Fintuning & 41.19 &   29.42   & 26.19& 38.96  \\ 
MiniLLM          & 32.20 &   23.64   & 21.31 & 30.24   \\ 
\rowcolor[HTML]{C0C0C0}KL Loss(Ours)  & 32.23 &   23.65 &21.29 & 30.22\\
\bottomrule
\end{tabular}
}
\label{tab5}
\end{table}

\subsection{Speedup}

Table \ref{tab6} presents speedup results obtained using TensorRT-LLM\footnote{https://github.com/NVIDIA/TensorRT-LLM}. We evaluate the actual end-to-end decoding speedup on two GPU architectures with the LLaMA2-7B 2:4 sparse model. We employ throughput, measured as the number of tokens processed per second, as the primary evaluation metric. Across a range of input and output lengths, the 2:4 sparse model demonstrates an overall acceleration of 1.33× to 1.83× compared to its dense counterpart, highlighting its potential for practical deployment.

\begin{table}[h]
\centering
\caption{Speedup results using TensorRT-LLM on RTX4090 and L20 GPUs with different input and output sequence lengths, measured by throughput (tokens/s). }
\resizebox{0.45\textwidth}{!}{%
\begin{tabular}{cccc}
\toprule 
\multicolumn{4}{c}{RTX4090} \\ \midrule
\multicolumn{1}{c|}{Inp Len, Out Len} & Sparse & Dense & Speedup \\ \midrule
\multicolumn{1}{c|}{128, 128} & 70.23 & 52.94 & 1.33x \\  
\multicolumn{1}{c|}{128, 1024} & 69.11 & 52.00 & 1.33x\\  
\multicolumn{1}{c|}{1024, 128} & 68.06 & 51.10 &  1.33x\\ 
\multicolumn{1}{c|}{1024, 1024} & 67.41 & 50.37 &  1.34x\\ \midrule
\multicolumn{4}{c}{L20} \\ \midrule
\multicolumn{1}{c|}{Inp Len, Out Len} & Sparse & Dense & Speedup \\ \midrule
\multicolumn{1}{c|}{128, 128} & 54.75 & 29.86 & 1.83x \\  
\multicolumn{1}{c|}{128, 1024} & 53.81 & 29.57 & 1.82x \\  
\multicolumn{1}{c|}{1024, 128} & 52.49 & 29.18 & 1.80x \\ 
\multicolumn{1}{c|}{1024, 1024} & 51.64 & 28.94 &  1.78x\\ 
 \bottomrule
 \label{tab6}
\end{tabular}
}

\end{table}

\section{Conclusion}
In this paper, we introduce the Adaptive Sparse Trainer (AST), a novel and efficient training pipeline for semi-structured sparse models. AST effectively narrows the precision gap between dense and sparse LLMs in terms of perplexity and accuracy on zero-shot tasks, while keeping training costs minimal. Our results demonstrate that pruning LLMs is feasible with minimal performance loss on knowledge-intensive tasks, and that large semi-structured sparse models can outperform dense models of similar decoding speed when supported by the appropriate software and hardware. Although our findings contribute to advancing the retraining of pruned models with billions of parameters, we acknowledge that the limited number of training tokens, due to computational constraints, remains an area for future exploration. Expanding this work to larger models or increasing the number of training tokens could provide valuable insights and further enhance the effectiveness of our methods.
\section*{Acknowledgements}
The authors would like to thank Ziteng Wang, Bingrui Li and Pengle Zhang for helpful discussions and suggestions on code implementation. This work was supported by NSFC Project (No.~62376131). 

\section{Appendix}
\subsection{Compressing Models with Different Semi-structured Sparsity Pattern}
In previous sections, we only examined 2:4 patterns, but theoretically, any N:M pattern can enhance matrix multiplication with appropriate hardware support. With that said, patterns with looser constraints (i.e., a larger M) often perform better but are more challenging to compress effectively. We aim to identify the optimal trade-off and. To maintain fairness in comparison, we focus only on patterns that achieve a 50\% sparsity ratio, such as 2:4, 4:8, etc.

It should be noted that deploying semi-structured sparse models can lead to potential memory overhead. Take the 2:4 sparsity pattern as an example: although only two non-zero elements out of every four weights need to be stored, additional positional information is required to correctly restore these compressed weights to the original matrix for calculations. Storing these indices will consume an additional proportion of memory. A naive method would be to store one bit as a mask for each element, which results in a 50\% memory overhead with a 4-bit quantized model at 50\% sparsity. However, a more efficient approach exists: since there are 2 non-zero elements, there are a total of ${4 \choose 2} = 6$ possible positions, hence only an additional 3 bits are sufficient.

We define the compression ratio as the percentage of memory latency of the compressed model compared to the original model. In our setting, we compress dense models with FP32 precision to a 4-bit model with $n$:$2n$ semi-structured sparsity; therefore, the compression ratio can be formulated as:
\begin{align}
    C_{n} = \frac{n*4 + \lceil \log_2{{2n \choose n}}\rceil (bit) } {2n*32(bit)} = \frac{1}{16} + \frac{\lceil \log_2{{2n \choose n}}\rceil}{64n}. \nonumber
\end{align}
Through mathematical formulation (provided in Section 2 of the Appendix), we can provide an approximate upper bound for the compression ratio as well as a predicted actual compression ratio when extending to sparsity patterns with larger $N$ and $M$
\begin{align}
     \log_2 {2n \choose n} = \log_2 \frac{(2n)!}{(n!)^2} \leq \log_2 \frac{ 4^{n}}{\sqrt{\pi n}} = 2n  - \log_2 \sqrt{\pi n}. \nonumber
\end{align}
If we remove the ceiling function for approximation:
\begin{align}
    C_n \approx \frac{3}{32} - \frac{\log_2 \sqrt{\pi n}}{64 n}.\nonumber
\end{align}
Therefore, the compression ratio is approximately an increasing function with an upper bound of $C^* = \frac{3}{32} \approx 9.375\%$. In practice, to further enhance compression, we can employ Huffman Encoding, similar to what was done in Deep Compression \cite{han2015deep}. We construct a Huffman Tree with evenly distributed frequencies, as empirical results support our claim. The actual compression ratios are shown in Table \ref{app:tab1}. We observe that as $n$ increases, performance improves; however, although the compression ratio increases, it remains below the upper bound.

\begin{table*}[t]
\caption{Comparison of perplexity and compression ratio of different sparsity pattern.}
\label{app:tab1}
\centering
\resizebox{0.8\textwidth}{!}{%
\begin{tabular}{c|c|c|c|c|c|c|c}
\toprule
\textbf{Sparsity Pattern} & \textbf{1:2} & \textbf{2:4} & \textbf{4:8} & \textbf{8:16} & \textbf{16:32} & \textbf{32:64} & \makecell{\textbf{Unstructured}\\ \textbf{50\%}} \\ \midrule 
 
Perplexity                       & 32.56  & 31.13  & 30.73  & 30.37  & 30.34  & 30.32  & 30.18 \\  \midrule
\makecell{Actual Compression\\ Ratio}                & 7.81\% & 8.33\% & 8.66\% & 8.93\% & 9.10\% & 9.22\% & -     \\ \bottomrule
\end{tabular}
}
\end{table*}

\subsection{Proof for Upper Bound of Combination Number} \label{A1}
Proofs are sourced from Stack Exchange \footnote{https://math.stackexchange.com/questions/1442351/stirling-approximation-of-binom2nn}. We have the follow equation:
\begin{align}
    \frac{1}{4^n} {2n \choose n} = \frac{(2n-1)!!}{(2n)!!} = \prod_{k=1}^{n} (1-\frac{1}{2k}). \nonumber
\end{align}
We square both sides of the equation:

\begin{align}
    (\frac{1}{4^n} {2n \choose n})^2 = \frac{1}{4} \prod_{k=2}^{n} (1-\frac{1}{2k})^2   = \frac{1}{4n} \prod_{k=1}^{n-1} (1-\frac{1}{(2k+1)^2})^{-1} .\nonumber
\end{align}

Using the Weierstrass product for the cosine function, we obtain:

\begin{align}
    \prod_{k=1}^{\infty} (1 - \frac{1}{(2k+1)^2})^{-1}   =\frac{4}{\pi}. \nonumber
\end{align}
Hence, it follows that:
\begin{align}
    (\frac{1}{4^n} {2n \choose n})^2 = \frac{1}{\pi n} \prod_{k\geq n}^{n} (1-\frac{1}{(2k+1)^2}) = \frac{1}{\pi n} \prod_{k\geq n}^{n} (1+\frac{1}{(2k+2)2k})^{-1}. \nonumber
\end{align}

Therefore we have

\begin{align}
    \frac{1}{\sqrt{\pi n}} \geq \frac{1}{4^n} {2n \choose n}. \nonumber
\end{align}

\subsection{Hyperparamters} \label{A2}
We present the hyper-parameters and the number of tokens used to train our model in Table \ref{hyper}. The increase iteration number $T_0$ is selected to be one-fourth of the total training iteration $T_1$.

\begin{table*}[t]
\centering
\caption{Summary of hyperparameters and the number of tokens used for training.}
\label{hyper}
\resizebox{0.8\textwidth}{!}{%
\begin{tabular}{c|ccc|cccc|c}
\toprule
& \multicolumn{3}{c|}{OPT} & \multicolumn{4}{c|}{GPT2}     & LLAMA2 \\ \midrule
& 125M    & 350M  & 1.3B   & 124M  & 350M  & 774M  & 1.5B  & 7B     \\
\rowcolor[HTML]{C0C0C0} 
Learning Rate      & 1e-4    &  1e-4   & 2e-5   & 2e-4  & 1e-4  & 1e-4  & 6e-5  & 2e-5   \\
Maximal Decay Factor  & 3e-4    &   1e-4    & 6e-5   & 1e-4  & 6e-5  & 6e-5  & 6e-5  & 6e-5   \\
\rowcolor[HTML]{C0C0C0} 
Training Steps      & 40k     & 40k   & 20k    & 40k   & 40k   & 40k   & 20k   & 15k    \\
Total Flipped Ratio & 11.1\%  &    9.3\%   & 4.4\%  & 6.7\% & 3.6\% & 3.4\% & 4.2\% & 3.8\%  \\
\rowcolor[HTML]{C0C0C0} 
Kl loss parameter     &   $2/3$   &  $2/3$  & $2/3$    & $2/3$  & $2/3$ &  $2/3$  &  $2/3$  &   $1/3$  \\
Tokens Trained      & 10B     & 10B    & 5B     & 5B    & 5B    & 5B    & 2.5B  & 7.5B  \\
\bottomrule
\end{tabular}%
}
\end{table*}

\subsection{SLoRB Initialization} \label{A4}
We conducted an ablation study on various initialization methods for SLoRB. We report the results on GPT2 model under specified computation limits of 1B, 2.5B, and 5B tokens, as shown in Table \ref{a4}. 'Mean' refers to the initialization method discussed in the paper, while 'Zero' indicates the initialization of both matrices from zero. Our initialization method assists in maintaining model performance at the outset, thereby accelerating convergence.

\begin{table}[h]
\centering
\caption{Results of perplexity using different initialization methods with fixed training tokens.}
\label{a4}
\centering
\begin{tabular}{llllll}
\toprule
\textbf{Trained Token} & \textbf{0} & \textbf{1B} & \textbf{2.5B} & \textbf{5B} &  \\ \midrule
Mean                   & 475.05     & 32.78       & 31.78         & 31.13       &  \\
Xavier Uniform         & 720.74    & 33.22       & 31.82         & 31.15       &  \\
Zero                   & 720.74     & 33.43       & 32.93         & 32.46       & \\ 
\bottomrule
\end{tabular}

\end{table}

\begin{table*}[h]
\centering
\caption{Wikitext perplexity for quantized sparse models.}
\label{awq}
\resizebox{0.8\textwidth}{!}{%
\begin{tabular}{ccccccccc}
\toprule
         &        & LLAMA2 & \multicolumn{2}{c}{OPT} & \multicolumn{4}{c}{GPT2}        \\ \cline{3-9} 
Method &
  \begin{tabular}[c]{@{}c@{}}Theoretical 
 \\ Compression Rate\end{tabular} &
  7B &
  125M &
  1.3B &
  124M &
  350M &
  774M &
  1.5B \\ \midrule
 Dense    & 1.0x   & 5.12   & 27.76       & 14.62     & 29.95  & 21.72  & 19.43 & 17.40 \\
\rowcolor[HTML]{C0C0C0} 
AWQ-4bit & 0.125x  &  5.68  & 29.12       & 14.95     & 31.93  & 22.66  & 19.89 & 17.80 \\ 
AWQ-2bit & 0.0675x &  2.2e5  & 216.11      & 53.42     & 751.15 & 453.45 & 70.36 & 46.17 \\
\rowcolor[HTML]{C0C0C0} \begin{tabular}[c]{@{}c@{}}AST-Naive-8bit\\ (Ours)\end{tabular} &
  0.125x &
    6.37     &
  30.26 &
  15.86 &
  32.36 &
  23.66 &
  21.29 &
  18.34 \\
\begin{tabular}[c]{@{}c@{}}AST-Naive-4bit\\ (Ours)\end{tabular} &
  0.0675x &
   6.48&
  31.28 &
  16.05 &
  34.00 &
  24.39 &
  21.72 &
  18.60 \\
\rowcolor[HTML]{C0C0C0}  \begin{tabular}[c]{@{}c@{}}AST-Boosted-4bit\\ (Ours)\end{tabular} &
  0.078x &6.25 &  29.88 &  15.63 &  32.76 &  23.87&  21.40& 18.31
  \\
  \bottomrule
\end{tabular}%
}
\end{table*}

\subsection{Different Distillation Function}
In this section, we formally present the loss functions applied in our ablation study, focusing on different distillation methods. We include results from previous works such as TinyBERT \cite{jiao2019tinybert}, MobileBERT \cite{sun2020mobilebert}, Sparse-Finetuning \cite{kurtic2023sparse}, and MiniLLM \cite{gu2024minillm}.

\subsubsection{TinyBERT}
In TinyBERT, we utilize information from intermediate multi-head attentions and hidden states using the mean square error (MSE) loss. Specifically, the attention loss is defined as:
\begin{align}
    \mathcal{L}_{attn} = \frac{1}{h} \sum_{i=1}^{h} \text{MSE}(A_i^S, A_i^T),
\end{align}
where $A_i^S$ and $A_i^T$ represent the layer-wise attention outputs of head $i$ from the student model and the teacher model, respectively.

The hidden state loss is defined as:
\begin{align} \label{app:eq2}
    \mathcal{L}_{hidn} =  \text{MSE}(H^S W_h, H^T),
\end{align}
where $H^S$ and $H^T$ represent the layer-wise hidden states of the student model and teacher model, respectively. Here, $W_h$ is a learnable projection matrix. However, since the shapes of the hidden states are identical in our settings, we fix $W_h = I$.

TinyBERT also uses cross-entropy loss:
\begin{align}
    \mathcal{L}_{pred} = \text{CE}(z^T/\tau, z^S/\tau),
\end{align}
where $z^T$ and $z^S$ are the output logits of the teacher and student models, respectively. We do not utilize embedding distillation, as the embeddings are well-trained and not pruned.

The distillation function for each layer is defined as:
\begin{align}
    \mathcal{L}_{layer} = \begin{cases}
        \mathcal{L}_{pred} & \text{if it is the last layer},\\
        \mathcal{L}_{hidn} + \mathcal{L}_{attn} & \text{otherwise}.
    \end{cases}
\end{align}

The final loss function is the sum of the layer-wise loss functions.

\subsubsection{MobileBERT}
MobileBERT is designed for distilling BERT by utilizing information from feature maps and attention. The feature map distillation in MobileBERT is identical to that used in TinyBERT, as shown in Equation \ref{app:eq2}. For attention knowledge transfer at layer $l$, the loss function is defined as:
\begin{align}
    \mathcal{L}_{AT}^{l} = \frac{1}{AT} \sum_{t=1}^{T} \sum_{a=1}^{A} D_{KL}(a_{t,l,a}^{tr} \| a_{t,l,a}^{st}),
\end{align}
where $A$ is the number of attention heads, $T$ is the sequence length, and $a_{t,l,a}$ represents the attention output for layer $l$, head $a$, and token $t$. The total distillation loss is the sum of the feature map loss and attention transfer.

\subsubsection{Sparse-Finetuning}
Sparse-Finetuning employs both SquareHead Loss and KL-divergence loss. It measures the intermediate representation using mean squared error (MSE):
\begin{align}
     \mathcal{L}_{feat} = \frac{\text{MSE}(f_t^l, f_s^l)}{\text{MSE}(f_t^l, 0)}, 
\end{align}
where $f_t^l$ and $f_s^l$ are the hidden states of the $l$-th layer at step $t$ for the teacher and student models, respectively.

KL-divergence loss \cite{kullback1951information} is used to measure the difference in output probability distributions between the student and teacher models:
\begin{align}
    \mathcal{L}_{logit} = D_{KL}(p_{\theta_t} \| p_{\theta_s}), 
\end{align}
where $p_{\theta_t}$ and $p_{\theta_s}$ are the probability distributions of the teacher and student models, respectively.

The total distillation loss is the weighted sum:
\begin{align}
    \mathcal{L} = \alpha_1 \mathcal{L}_{logit} + \alpha_2 \mathcal{L}_{feat}.
\end{align}

\subsubsection{MiniLLM}
MiniLLM uses reverse KL-divergence instead of forward KL. Specifically, the loss function is defined as:
\begin{align}
    \mathcal{L}_{logit} = D_{KL}(p_{\theta_s} \| p_{\theta_t}),
\end{align}
where $p_{\theta_t}$ and $p_{\theta_s}$ are the probability distributions of the teacher and student models, respectively.

\subsection{Adding Decay for AdamW Optimizer}
Recent work \cite{hu2024accelerating} has shown that applying SR-STE with momentum-based optimizers can weaken the effectiveness of the decay term, resulting in suboptimal performance. In earlier implementations, SR-STE added the decay term directly to the weight. Specifically, we denote the gradient with respect to weight $W_t$ at iteration $t$ as $g_t$ which is further adjusted using the AdamW optimizer, followed by the weight update using:
\begin{align}
    \text{AdamW}(g_t) = \frac{ (u_t \beta_1 + (1-\beta_1) g_t)}{(1-\beta_{1}^{t})(\sqrt{v_t} + \epsilon)}
\end{align}

\begin{align}
    W_{t+1} \leftarrow W_{t} - \gamma_t \left( \text{AdamW}(g_t) + \lambda_W (\overline{m(W_t)} \odot W_t) \right).
\end{align}
where $u_t$ and $v_t$ are the first-order and second-order momenta of $g_t$, respectively. $\gamma_t$ is the current learning rate.

However, results indicate that this approach causes frequent mask oscillations details can be found in previous work \cite{hu2024accelerating}. To mitigate this issue, the decay term is instead applied directly to the weight using the following update rule:
\begin{align}
    \tilde{g_{t}} \leftarrow g_{t} + \lambda_W (\overline{m(W_t)} \odot W_t),
\end{align}
\begin{align}
    W_{t+1} \leftarrow W_{t} - \gamma_t \text{AdamW}(\tilde{g_{t}}),
\end{align}

We make further improvement to this method, since the decay term is only dependent to the parameter's current value it should not be entangled with first order momentum. Therefore we employ the following update role:

\begin{align}
    u_{t} = u_{t-1} \beta_1 + (1-\beta_1) g_{t-1}
\end{align}

\begin{align}
    \tilde{u_{t}} =u_{t} + \lambda_W (\overline{m(W_t)} \odot W_t)
\end{align}

Furthermore we use the decayed gradient to calculate the second-order momentum and update weights accordingly.

\begin{align}
    v_{t} = v_{t-1} \beta_2 + (1-\beta_2) \tilde{u_{t-1}}^2
\end{align}

\begin{align}
    W_{t+1} \leftarrow W_{t} - \gamma_t \frac{\tilde{u_{t}} }{(1-\beta_{1}^{t})(\sqrt{v_t} + \epsilon)}.
\end{align}

\subsection{Mask Oscillation Behavior During Training} \label{A5}

Several metrics have been introduced to measure the stability of masks during training, such as flip rate \cite{hu2024accelerating} and SAD \cite{zhou2021learning}. In this paper, we primarily focus on measuring the flip rate and initial flip rate, which quantify the mask changes between the current and previous steps, as well as between the current and initial steps, respectively.
\begin{align}
     r_t = ||m(W_{t}) - m(W_{t-1})||_1/D \nonumber
\end{align}
\begin{align}
     i_t= ||m(W_{t}) - m(W_{0})||_1/D \nonumber
\end{align}
where $D$ represents the total number of parameters in the linear layer. The flip rate reflects the stability of the model masks, while the initial flip rate indicates the overall extent of weight removal and revival.

\begin{figure}[h]
    \centering

    \begin{subfigure}
        \centering
        \includegraphics[width=0.32\textwidth]{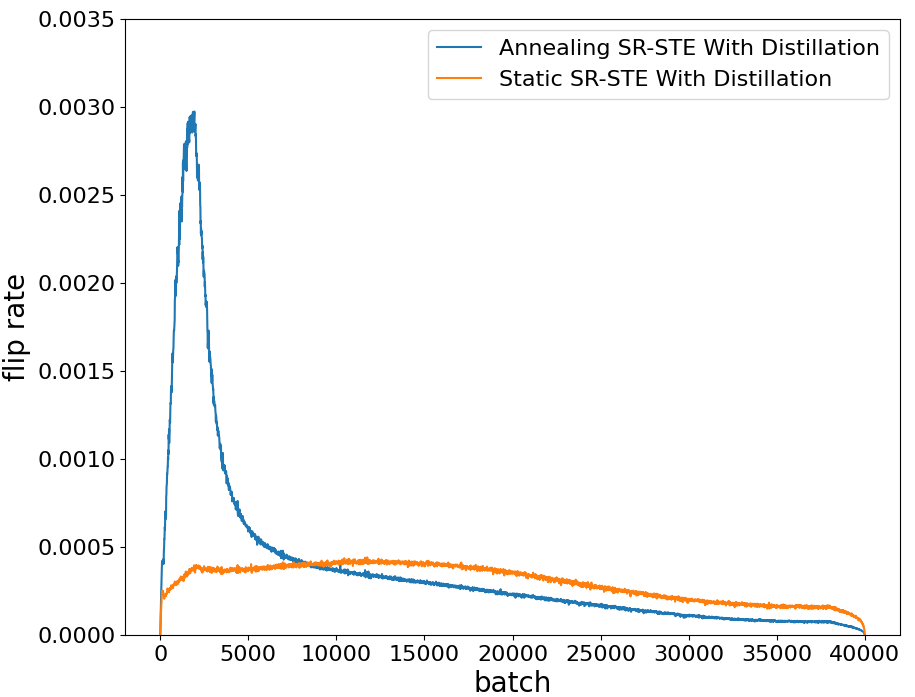}
        \label{pic4:sub1}
    \end{subfigure}

    \begin{subfigure}
        \centering
        \includegraphics[width=0.32\textwidth]{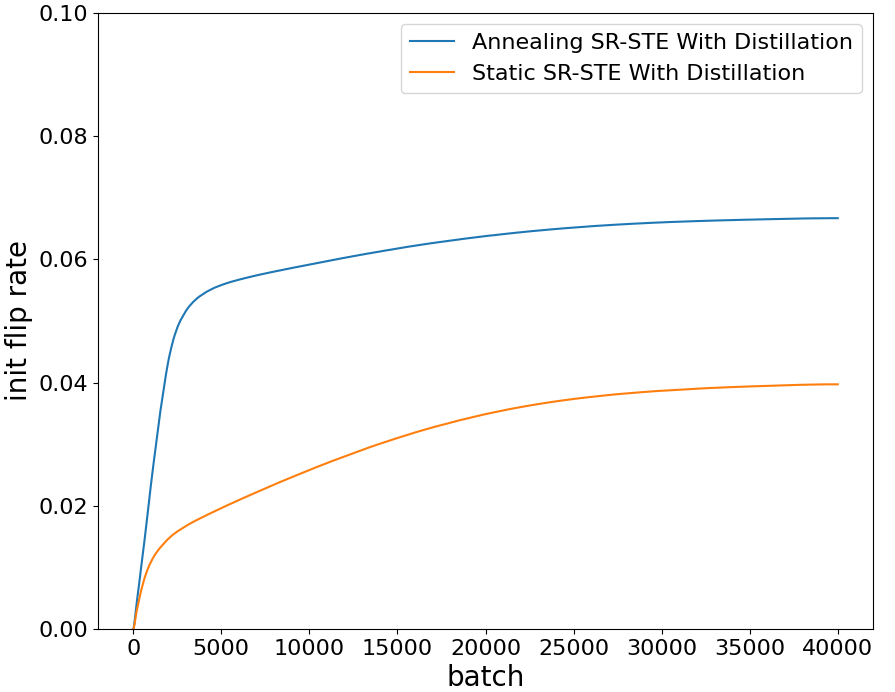}
        \label{pic4:sub2}
    \end{subfigure}

    \caption{\textbf{(Upper)} Flip rate for static and Annealing SR-STE during the training of the GPT2 model.  \textbf{(Lower)} Initial flip rate for static and Annealing SR-STE during the training of the GPT2 model.}
    \label{pic4}
\end{figure}

We compare the flip rates and initial flip rates of static SR-STE and Annealing SR-STE. In our experiments, the mask is updated, and the flip rate and initial flip rate are calculated every 10 batches during training, as more frequent recalculations do not improve accuracy and only increase computational overhead. Figure \ref{pic4} illustrates the flip rates and initial flip rates during the retraining of the GPT2 model. Unlike traditional static decay factors, Annealing SR-STE modifies a higher percentage of the mask at the beginning, allowing the model to explore various mask patterns. Additionally, Annealing SR-STE enables a more stable mask towards the end of training, promoting better convergence. As a result, Annealing SR-STE supports a higher rate of mask changes (e.g., initial flip rate) while maintaining overall stability.

\subsection{Model Compression Using AWQ} \label{A3}

In this section, we present the perplexity results for quantized models using AWQ. As shown in Table \ref{awq}, our model maintains its performance even under extreme compression. For the AST-Boosted model, we also quantized the additional SLoRB parameters.

\subsection{Frequently Asked Questions}

This section examines and addresses key concerns raised in prior discussions.

\subsubsection{Justification of Knowledge Distillation}
One concern is that the use of distillation may introduce additional computational costs. However, we have found that, under the same total FLOPs, distillation methods still outperform non-distillation methods. We present a performance comparison under identical computational costs, and the results in Table \ref{KD} demonstrate that the distilled model significantly outperforms the non-distilled model.
\begin{table}[h]
\centering
\caption{Perplexity for distillation and distillation-free methods under the same computational cost.}
\label{KD}
\resizebox{0.5\textwidth}{!}{%
\begin{tabular}{cccc}
\toprule
\textbf{Module} &  \textbf{Tokens} & \textbf{Total(TFLOPs)} & \textbf{PPL} \\ \hline
 LLaMA2 w/ Distil &  5B &  3e8 & 5.97 \\ \hline
 LLaMA2 w/o Distil &  7.5B &  3e8 & 6.12   \\
  \bottomrule
\end{tabular}
}
\end{table}

\subsubsection{Justification for SLoRB}
One concern is the timing of training the additional SLoRB component. Through experimentation, we observed that fine-tuning the LoRA component after initial training did not yield further performance improvements. We hypothesize that this is because the model has already converged on the pretraining dataset, making it difficult to further enhance performance through fine-tuning the LoRA module. In contrast, keeping both the model and LoRA modules unfrozen during retraining offers greater flexibility and potential for improvement. To support this hypothesis, we present the perplexity results for the GPT2 model in Table \ref{SLoRB}."

\begin{table}[h]
\centering
\caption{SLoRB ablation results on GPT2.}
\label{SLoRB}
\resizebox{0.5\textwidth}{!}{%
\begin{tabular}{c|cccc}
\toprule
\textbf{Model} & \textbf{SLoRB} & \textbf{w/o LoRA} & \textbf{Finetune Aft. Retrain} & \textbf{Dense} \\ \hline
GPT2 & \textbf{31.13} & 32.23 & 32.07 & 29.95 \\
  \bottomrule
\end{tabular}
}
\end{table}

\subsubsection{Mask Selection}
We present computational complexity and perplexity results for different mask selection criteria, such as the Wanda and SparseGPT metrics, in Table \ref{Mask}. The results demonstrate that the magnitude metric outperforms the others in both performance and speed.

\begin{table}[h]
\centering
\caption{Wikitext perplexity for quantized sparse models.}
\label{Mask}
\resizebox{0.35\textwidth}{!}{%
\begin{tabular}{c|cc}
\toprule
\textbf{Metric} & \textbf{PPL} & \textbf{Complexity} \\ \hline
Magnitude & 32.23 & O($h^2$) \\ \hline
SparseGPT & 32.58 & O($h^3$) \\ \hline
Wanda & 32.51 & O($h^2$) \\
  \bottomrule
\end{tabular}
}
\end{table}

\subsubsection{Effects of $\alpha$}
We provide results using different methods on full GPT2 retraining and LLaMA2 retraining with limited training tokens in Table \ref{alpha}. When $\alpha = 1$, meaning we only use KD loss, we observe slightly better performance on smaller models. However, it performs less effectively on larger models like LLaMA2-7B, where tuning $\alpha$ leads to improvements in performance. This suggests that $\alpha$ plays a more critical role in optimizing results for larger-scale models.

 \begin{table}[h]
\centering
\caption{Wikitext perplexity for different $\alpha$ on LLaMA2-7B and GPT2 model.}
\label{alpha}
\resizebox{0.49\textwidth}{!}{%
\begin{tabular}{c|ccccc}
\toprule
\textbf{Model} & \textbf{Tokens Trained} & $\alpha=0$ & $\alpha=\frac{1}{3}$ & $\alpha=\frac{2}{3}$ &  $\alpha=1$\\ \hline
LLaMA2-7B & 3B & 6.26 & 6.25 & \textbf{6.22} & 6.62 \\
GPT2 & 5B & \textbf{32.18} & 32.23 & 33.57 & 40.34  \\
  \bottomrule
\end{tabular}
}
\end{table}

\bibliography{aaai25}

\end{document}